\documentclass[10pt,twocolumn,letterpaper]{article}

\usepackage[pagenumbers]{paperstyle}

\usepackage{multirow}
\usepackage{xspace}
\usepackage{amsmath,amssymb}
\usepackage{booktabs}
\usepackage{colortbl}
\usepackage{caption}
\usepackage{microtype}
\usepackage{cuted}

\definecolor{linkblue}{rgb}{0.21,0.49,0.74}
\definecolor{ourbg}{RGB}{226,238,248}
\usepackage[pagebackref,breaklinks,colorlinks,allcolors=linkblue]{hyperref}

\newcommand{\ours}{FieldGuideRAG\xspace}
\newcommand{\qa}{Qwen2.5-VL-7B\xspace}
\newcommand{\qb}{Qwen3-VL-8B\xspace}
\newcommand{\qc}{Qwen3-VL-32B\xspace}
\newcommand{\qas}{Qwen2.5\xspace}
\newcommand{\qbs}{Qwen3\xspace}
\DeclareMathOperator*{\argmax}{arg\,max}
\newcommand{\gr}[1]{{\color{gray}#1}}
\newcommand{\best}[1]{\textbf{#1}}
\newcommand{\ourrow}{\rowcolor{ourbg}}

\title{FieldGuideRAG: Image-Grounded GraphRAG for Fine-Grained\\Recognition with Vision-Language Models}

\author{
Jiacheng Ding\textsuperscript{1}\quad
Yanliang Qi\textsuperscript{1}\quad
Cong Guo\textsuperscript{1}\quad
Peihua Mai\textsuperscript{2,$\ast$}\\
\textsuperscript{1}University of Memphis\quad
\textsuperscript{2}National University of Singapore\\
{\tt\small \{jding2, yqi2, cguo\}@memphis.edu}\quad
{\tt\small peihua.m@u.nus.edu}\\
{\small $^{\ast}$Corresponding author}
}

\hypersetup{pdftitle={FieldGuideRAG: Image-Grounded GraphRAG for Fine-Grained Recognition with Vision-Language Models},pdfauthor={Jiacheng Ding, Yanliang Qi, Cong Guo, Peihua Mai}}
\begin{document}
\maketitle
\begin{strip}
\centering
\includegraphics[width=\textwidth]{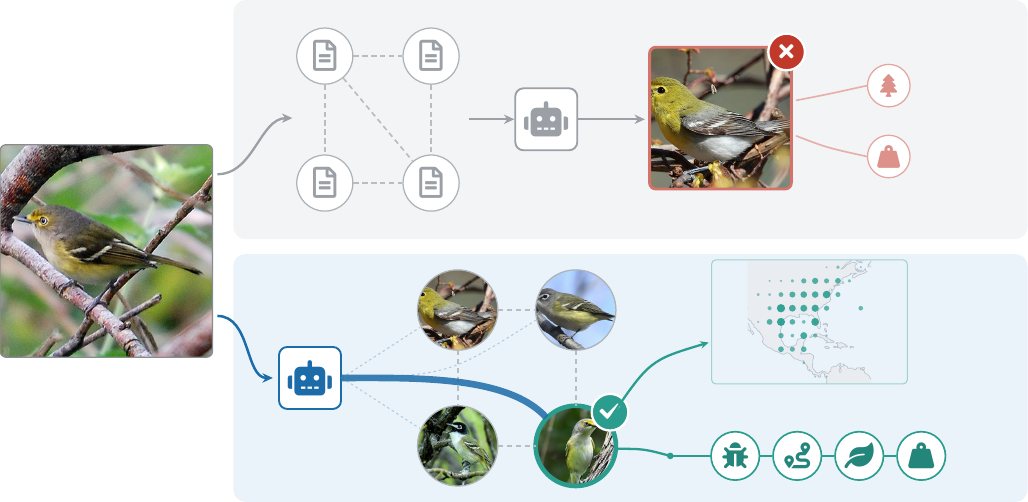}
\captionsetup{hypcap=false}
\captionof{figure}{\textbf{\ours.} A GraphRAG with text nodes leads \qb to a look-alike and to the facts of that look-alike in the upper row, and our image-grounded graph lets the VLM verify the query against each node photograph and follow the range and fact edges of the verified species in the lower row.}
\captionsetup{hypcap=true}
\label{fig:teaser}
\end{strip}

\begin{abstract}
We present \ours, a training-free GraphRAG method in which a vision-language model, or VLM, recognizes fine-grained species by comparing the query with real photographs stored on the nodes of a knowledge graph. Multimodal GraphRAG systems describe each entity in text, and text conveys little of the visual detail that separates look-alike species. Our graph stores one photograph on every species node and links each species to its relatives, its geographic range and its ecological facts. A frozen VLM verifies the query against the photograph of each candidate node. \ours fuses the verification scores with the range edge and answers questions about the organism from the fact edges of the verified species. On CUB-200, a Wikidata graph and four iNaturalist domains, the node photograph improves verification over the species name on every graph, while text descriptions add little. Verification of the candidate pool exceeds a SigLIP retriever by up to 16 points, the range edge adds 7 to 19 points on iNaturalist, and answering from fact edges is correct on up to 89.5\% of 28{,}820 ecological questions. The verification scores also detect queries whose species is missing from the graph. Graphs, prompts and code will be released.
\end{abstract}

\section{Introduction}
\label{sec:intro}

Vision-language models, or VLMs, describe everyday scenes well and often fail when categories differ in small visual details. On fungi and Lepidoptera from iNaturalist, the VLMs we test rarely name the correct species without help. Retrieval-augmented generation lets a model consult external knowledge at inference time~\cite{lewis2020rag}. GraphRAG organizes that knowledge as entities and typed relations, which supports multi-hop questions and makes the retrieved evidence traceable~\cite{edge2024graphrag,gutierrez2024hipporag,he2024gretriever,guo2024lightrag}. We ask what a knowledge graph should give a VLM when the answer depends on fine visual differences.

Current systems give the VLM text about each entity. Knowledge-based VQA systems link the query image to an encyclopedia entity and read its article~\cite{caffagni2024wikillava,yan2024echosight}. Multimodal GraphRAG systems such as mKG-RAG, MMGraphRAG, M3KG-RAG and MegaRAG build graphs from images, videos and documents, and they describe each entity in text~\cite{yuan2025mkgrag,wan2025mmgraphrag,park2026m3kgrag,hsiao2026megarag}. None of these systems lets the answering model compare the query photograph with a reference photograph of each candidate entity. Field marks such as the completeness of an eye ring or the streaking of a breast are hard to put into words. Multi-image RAG places retrieved photographs in one prompt~\cite{hu2025mragbench}. The VLM must then compare many images at once, and VLMs remain weak at such comparisons~\cite{fu2024blink,wang2025muirbench}. A multi-image prompt also lacks the relations that lead from a recognized species to further knowledge.

We present \ours, which follows the design of a printed field guide. A field guide shows the photograph of each species beside its look-alikes, its range map and notes on its habits. Our image-grounded knowledge graph stores one real photograph on every species node and connects each species to its siblings, to the cells of its geographic range and to its ecological facts. A SigLIP retriever finds the entry nodes~\cite{zhai2023siglip}, and sibling edges add the look-alikes of the most similar species to the candidate pool. A frozen VLM then compares the query with the photograph of each candidate in a separate prompt and returns the probability that both show the same species. \ours fuses this probability with the range edge of each candidate and answers questions about the organism from the fact edges of the verified species. \Cref{fig:teaser} contrasts this process with a GraphRAG whose nodes hold text. The method needs no training, and its graphs come from public catalogues~\cite{vrandecic2014wikidata,tobias2022avonet,vanhorn2021inat}.

We evaluate \ours on CUB-200, on a graph built from Wikidata photographs and Wikipedia text and on four iNaturalist domains with two frozen Qwen-VL models. The node photograph helps the VLM on every graph, and Wikipedia text or expert attributes add little over the species name. Verification of the candidate pool is more accurate than the retriever on every graph. On the domains that the VLM rarely names from memory, verification is also more accurate than a multi-image prompt. The range and fact edges add information that embedding similarity lacks, and the verification scores detect queries whose species is absent from the graph.

We make three contributions. We build image-grounded knowledge graphs of species from public catalogues, in which every node holds a real photograph and typed edges hold taxonomy, range and facts. We propose \ours, a training-free method that verifies candidate nodes against their photographs and then follows the range and fact edges of the verified species. We measure under matched controls on seven graphs which node evidence, reading strategy and relations help a VLM recognize fine-grained species. Graphs, prompts and code will be released.

\section{Related Work}
\label{sec:related}

\paragraph{Retrieving knowledge for a VLM.}
Retrieval-augmented VLMs retrieve image and text pairs for captioning and question answering~\cite{yasunaga2023racm3,chen2022murag,hu2023reveal}. Knowledge-intensive benchmarks~\cite{hu2023oven,chen2023infoseek,mensink2023evqa} led to systems that link an image to a Wikipedia entity and read its article~\cite{caffagni2024wikillava,yan2024echosight,qi2024rora}. VisRAG retrieves page images in place of parsed text~\cite{yu2025visrag}. These systems return documents about an entity, and our graph returns a photograph of each candidate.

\paragraph{Structuring knowledge as a graph.}
GraphRAG and its variants retrieve entities, neighbourhoods and summaries from a graph that a language model builds from text~\cite{edge2024graphrag,guo2024lightrag,gutierrez2024hipporag,he2024gretriever}. mKG-RAG builds a multimodal graph for knowledge-intensive VQA and stores each entity as a name with a description~\cite{yuan2025mkgrag}. MMGraphRAG converts images into scene graphs and links them to a text graph~\cite{wan2025mmgraphrag}. M3KG-RAG links audio and video clips to triplets and passes text triplets to the answering model~\cite{park2026m3kgrag}. MegaRAG embeds figure entities through their names and descriptions and passes retrieved pages to the answering model~\cite{hsiao2026megarag}. Our graph stores a photograph on every node, and the VLM compares the query with that photograph.

\paragraph{Representing a category.}
Class descriptions written by language models improve zero-shot CLIP classification~\cite{pratt2023cupl,menon2023description}, and attribute annotations have long supported fine-grained recognition~\cite{wah2011cub}. Retrieval-augmented classifiers retrieve labelled images to support the prediction~\cite{long2022rac,lee2025ratta}. MRAG-Bench finds that VLMs gain more from retrieved images than from retrieved text~\cite{hu2025mragbench}. We compare the species name, its text and its photograph as node evidence under one verification protocol.

\paragraph{Context beyond appearance.}
Location priors improve species recognition~\cite{berg2014birdsnap,macaodha2019presence,chu2019geoaware,cole2023sinr}. BioCLIP trains a CLIP model on images from across the tree of life with their taxonomic names as text~\cite{stevens2024bioclip}. We store a count-based range model, taxonomic siblings and ecological facts as graph edges and combine them with the VLM score at inference time.

\paragraph{Comparing images with a VLM.}
VLMs remain weak at tasks that require comparing several images~\cite{fu2024blink,wang2025muirbench}, and language models attend unevenly to positions in long contexts~\cite{liu2024lostmiddle}. Agentic methods train VLMs to request visual evidence while they reason~\cite{openai2025thinking,zheng2025deepeyes}. Our verifier asks one yes-or-no question about two images per candidate and needs no training.

\section{Method}
\label{sec:method}

\begin{figure*}[t]
  \centering
  \includegraphics[width=\textwidth]{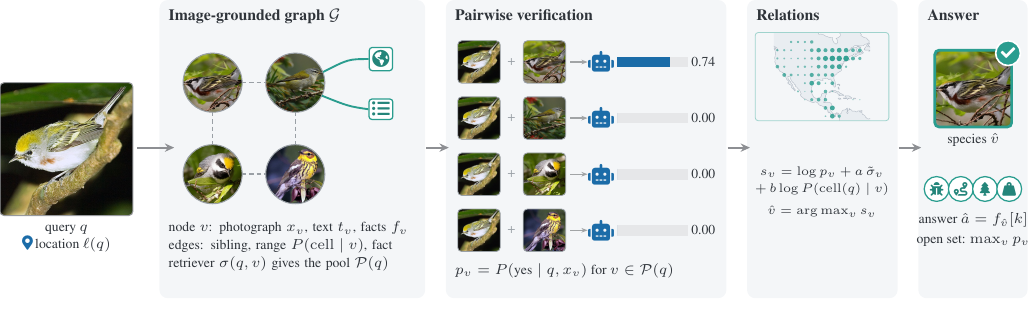}
  \caption{\textbf{Pipeline of \ours.} The retriever and the sibling edges select the candidate pool $\mathcal{P}(q)$ in the image-grounded graph, a frozen VLM verifies the query against the photograph of each candidate, the verification scores are fused with the range edge, and the fact edges of the verified species answer questions about the organism.}
  \label{fig:method}
\end{figure*}

\subsection{Problem definition and notation}
\label{sec:problem}
A knowledge graph $\mathcal{G}=(\mathcal{V},\mathcal{E})$ holds one node $v\in\mathcal{V}$ for each species. A node carries the species name, one photograph $x_v$, an optional text $t_v$ and an optional fact record $f_v$. The entry $f_v[k]$ of the record holds the answer to a question of type $k$, such as the diet of the species. Typed edges in $\mathcal{E}$ connect a species to its siblings, to the cells of its geographic range and to its facts. A query is a photograph $q$ with an optional location $\ell(q)$. Species recognition predicts the node $\hat v$ whose species the query shows. Fact answering returns $\hat a=f_{\hat v}[k]$ for a question of type $k$ about the photographed organism. Open-set detection decides whether the species of $q$ has a node in $\mathcal{G}$. A retriever gives a similarity $\sigma(q,v)$ between the query and each node, and a frozen VLM answers prompts that hold images and text. No component is trained, and a development split chooses the few weights of the method. \Cref{fig:method} shows the pipeline of \ours.

\subsection{Image-grounded knowledge graph}
\label{sec:graph}
We call $x_v$ the node photograph. On CUB and iNaturalist, the node photograph is one training or validation image of the species drawn at random. On the Wikidata graph, the node photograph is the image that Wikidata attaches to the species through property P18~\cite{vrandecic2014wikidata}. The node text is the description section of the English Wikipedia article. On CUB, the node text is a sentence generated from the human-annotated attributes of the species~\cite{wah2011cub}. The fact record of a bird species holds its diet, migration, habitat, foraging lifestyle and body mass from AVONET~\cite{tobias2022avonet}.

Sibling edges connect species of the same genus. CUB has no genus labels, and its sibling edges connect species that share a common-name group such as sparrows or warblers. On iNaturalist the graph also records the family of each species. A range edge links a species to each cell of a 5$^\circ$ grid of latitude and longitude in which the species was photographed in the iNaturalist 2021 mini training set~\cite{vanhorn2021inat}. That set holds about 50 geotagged photographs per species and none of our test photographs. Counts are smoothed over the 3$\times$3 neighbourhood of each cell and with an additive constant, which gives the likelihood $P(\text{cell}\mid v)$. Fact edges connect each CUB-200 species to its AVONET values. Every graph comes from public data without labelling of our own.

\subsection{Candidate pool}
\label{sec:pool}
A SigLIP so400m encoder~\cite{zhai2023siglip} embeds the query and every node photograph, and $\sigma(q,v)$ is their cosine similarity. The graph shortlist holds four species, starting with the most similar species and adding its most similar siblings, then members of its family and then the next most similar species. The similarity shortlist of size $K$ holds the $K$ most similar species. The candidate pool $\mathcal{P}(q)$ is the union of the eight most similar species, the siblings of the most similar species and the graph shortlist. On iNaturalist the pool also includes the species of the same family among the 30 most similar ones.

\subsection{Pairwise verification}
\label{sec:verify}
For each candidate $v\in\mathcal{P}(q)$, the VLM receives the query, the node photograph $x_v$ and the name of $v$. The prompt asks whether the organism in the query is the same species as the organism in the node photograph and requests a yes or no answer. The score $p_v$ is the probability of yes under a softmax over two logits at the first generated position, the largest logit among the tokens for yes and the largest among the tokens for no. Scoring one candidate takes one forward pass with two images and no decoding, and the passes for different candidates are independent and can run in parallel. Verification alone predicts the candidate with the largest $p_v$.

A sibling edge can carry a contrastive note, and the prompt of the candidate appends the note before the question. The look-alikes of a candidate are the other species of its graph shortlist. The note names up to three CUB attributes whose annotation frequency for the candidate exceeds the mean over its look-alikes by at least 25 percentage points. The note also names up to two attributes that fall short of that mean by the same margin.

\begin{figure*}[t]
  \centering
  \includegraphics[width=\textwidth]{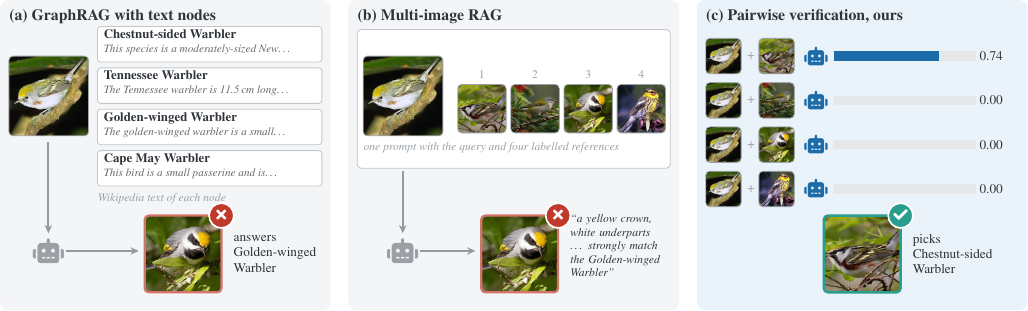}
  \caption{\textbf{Three ways to read a graph.} For the same Chestnut-sided Warbler query and the same four candidates, \qb picks a look-alike from text nodes and from one prompt with all four photographs, and pairwise verification gives the true species the only high score.}
  \label{fig:compare}
\end{figure*}

\Cref{fig:compare} contrasts pairwise verification with two other ways of giving a graph to a VLM. A GraphRAG with text nodes gives the VLM the description of each candidate, and the Wikipedia descriptions of the four warblers open with size and general shape. A multi-image prompt shows all candidate photographs at once, and the VLM must compare five photographs and choose one name. Pairwise verification asks one question about two photographs for each candidate. Every candidate receives its own score, and the scores combine with typed relations and stay comparable across queries.

\subsection{Fusion with the range edge}
\label{sec:fusion}
When the query has a location, \ours treats the verification score and the location as independent evidence about the species and scores each candidate by
\begin{equation}
  s_v=\log p_v + a\,\tilde{\sigma}_v + b\,\log P(\text{cell}(q)\mid v).
  \label{eq:fusion}
\end{equation}
Here $\text{cell}(q)$ is the grid cell that contains $\ell(q)$, $\tilde\sigma_v$ is $\sigma(q,v)$ standardized within the pool, and a grid search on the development split of each domain chooses the weights $a$ and $b$. \ours predicts $\hat v=\argmax_{v\in\mathcal{P}(q)} s_v$. Bayes' rule motivates this form. Let $p_v$ be the posterior probability of $v$ given the photograph, and let the location be independent of the photograph given the species. Then $\log P(v\mid q,\ell(q))$ equals $\log p_v+\log P(\text{cell}(q)\mid v)$ up to a term that does not depend on $v$. The weights $a$ and $b$ relax both assumptions, since $p_v$ answers a pairwise question and the retrieval similarity carries further evidence about the photograph. The development split also preferred this log-linear form over an additive combination of standardized scores.

\subsection{Fact answering and open-set detection}
\label{sec:hop}
A question of type $k$ about the photographed organism receives the answer $f_{\hat v}[k]$ from the fact edge of the predicted species. CUB-200 has no locations, and its species is predicted by $p_v+\lambda\tilde\sigma_v$. The development split chooses $\lambda$ together with the candidate set among the graph shortlist, similarity shortlists and the whole pool. The largest $p_v$ over the pool scores whether the species of the query has a node in the graph.

\section{Experiments}
\label{sec:exp}

\begin{table}[t]
\centering\footnotesize
\setlength{\tabcolsep}{1.9pt}
\caption{\textbf{Graphs.} Number of species, test queries and development queries, source of the node text, mean size and recall of the candidate pool, and recall of the four-species graph shortlist on the test queries.}
\label{tab:data}
\begin{tabular}{lrrrlrrr}
\toprule
 & & & & & \multicolumn{2}{c}{pool} & shortlist \\
\cmidrule(lr){6-7}\cmidrule(lr){8-8}
Graph & species & test & dev & node text & size & recall & recall \\
\midrule
CUB-93 & 93 & 372 & 279 & attributes & 16.8 & 99.2 & 75.8 \\
CUB-200~\cite{wah2011cub} & 200 & 5{,}794 & 600 & attributes & 12.8 & 95.1 & 78.6 \\
Wikidata graph & 93 & 372 & 279 & Wikipedia & 16.4 & 99.5 & 90.1 \\
iNat birds~\cite{vanhorn2021inat} & 406 & 812 & 812 & -- & 14.3 & 83.4 & 54.6 \\
iNat fungi & 341 & 682 & 682 & Wikipedia & 10.8 & 71.6 & 52.2 \\
iNat Lepidoptera & 402 & 804 & 804 & Wikipedia & 18.7 & 71.5 & 39.3 \\
iNat plants & 401 & 802 & 802 & -- & 20.0 & 91.0 & 62.3 \\
\bottomrule
\end{tabular}
\end{table}

\subsection{Setup}
\label{sec:setup}
\paragraph{Graphs.}
\Cref{tab:data} lists seven graphs with one photograph per node. CUB-93 contains the 93 CUB species of the nine common-name groups with at least five species, such as sparrows, warblers and gulls. Its test set holds four images per species, and its development set holds three other test-split images per species. CUB-200 uses all 200 species and all test images, and its development set holds three training images per species that are not node photographs. The Wikidata graph has the species and queries of CUB-93 and takes its node photographs from Wikidata and its node texts from Wikipedia. The four iNaturalist graphs come from the 2021 validation set~\cite{vanhorn2021inat}. iNat fungi contains all fungi of that set. iNat birds, Lepidoptera and plants contain whole genera with at least three species until about 400 species are reached. Each iNaturalist species gives one node photograph, two test queries and two development queries, and every iNaturalist query carries GPS coordinates. A few fungi and Lepidoptera have no Wikipedia description, and their node text states that no description is available.

\paragraph{Models.}
We use \qa~\cite{bai2025qwen25vl} and \qb~\cite{qwen3vl2025}, which the tables abbreviate as \qas and \qbs. All models are frozen and run in 16-bit floating point, and generative baselines decode greedily. CUB queries and node photographs are cropped to the annotated bounding box of the bird.

\paragraph{Baselines.}
The retriever predicts the most similar species. The zero-shot baseline asks the VLM to choose a species from the full list of names in the graph. Text RAG gives the VLM the names and node texts of the four graph-shortlist candidates in one prompt. Multi-image RAG gives the VLM the names and node photographs of a shortlist in one prompt, in the order of retrieval similarity, and asks for the species. Both RAG baselines may answer with any species in the list.

\paragraph{Metrics.}
We report top-1 species accuracy on the test queries and AUROC for open-set detection. Paired comparisons use McNemar's exact test~\cite{mcnemar1947}. No weight is fitted on test queries.

\subsection{Main results}
\label{sec:main}
\begin{table*}[t]
\centering\small
\setlength{\tabcolsep}{3.7pt}
\caption{\textbf{Main results.} Species accuracy in percent with one photograph per node and open-set AUROC for queries whose species was removed from the graph, where shortlist methods see the four candidates of the graph shortlist, the range edge uses the GPS coordinates of the iNaturalist photographs and a dagger marks the only pool result without a significant McNemar difference from the retriever.}
\label{tab:main}
\begin{tabular}{llcccccc}
\toprule
& Method & CUB-200 & Wikidata graph & iNat birds & iNat fungi & iNat Lepidoptera & iNat plants \\
\midrule
& SigLIP retriever & 53.2 & 63.7 & 25.0 & 30.2 & 16.3 & 32.5 \\
\midrule
 & Zero-shot VLM & 51.9 & 45.4 & 25.5 & 4.8 & 5.5 & 14.5 \\
 & Text RAG, $K{=}4$ & 62.1 & 58.9 & -- & 29.6 & 14.9 & -- \\
 & Multi-image RAG, $K{=}4$ & 66.7 & \best{72.8} & 33.1 & 28.3 & 18.5 & 33.0 \\
 & Verification, $K{=}4$ & 63.0 & \best{72.8} & 30.8 & 33.1 & 20.5 & 36.0 \\
\ourrow  & \ours, pool & \best{69.0} & 70.7 & 39.4 & 34.5 & 28.4 & 38.8 \\
\ourrow \multirow{-6}{*}{\rotatebox{90}{\qa}} & \ours, pool, range edge & -- & -- & \best{58.5} & \best{41.8} & \best{44.4} & \best{57.5} \\
\midrule
 & Zero-shot VLM & 47.0 & 36.0 & 20.4 & 8.2 & 8.5 & 14.5 \\
 & Text RAG, $K{=}4$ & 62.4 & 60.8 & -- & 29.0 & 15.4 & -- \\
 & Multi-image RAG, $K{=}4$ & 65.7 & 65.9 & 35.2 & 32.8 & 23.3 & 37.2 \\
 & Verification, $K{=}4$ & 62.9 & \best{73.4} & 32.0 & 34.2 & 23.6 & 38.7 \\
\ourrow  & \ours, pool & \best{68.8} & 69.4$^{\dagger}$ & 38.7 & 38.1 & 32.6 & 42.6 \\
\ourrow \multirow{-6}{*}{\rotatebox{90}{\qb}} & \ours, pool, range edge & -- & -- & \best{57.6} & \best{46.2} & \best{44.9} & \best{58.2} \\
\midrule
 & Retriever, largest similarity & 0.663 & 0.642 & 0.570 & 0.583 & 0.541 & 0.583 \\
\ourrow  & \ours, largest $p_v$, \qa & \best{0.819} & \best{0.785} & 0.619 & 0.626 & 0.595 & 0.610 \\
\ourrow \multirow{-3}{*}{\rotatebox{90}{\scriptsize AUROC}} & \ours, largest $p_v$, \qb & 0.807 & 0.763 & \best{0.648} & \best{0.691} & \best{0.646} & \best{0.643} \\
\bottomrule
\end{tabular}
\end{table*}

\Cref{tab:main} compares \ours with the baselines on six graphs. The VLMs alone name many CUB birds and rarely name iNaturalist fungi or Lepidoptera. Text RAG improves on the zero-shot VLM, and on CUB-200 and the Wikidata graph it stays below both methods that read the node photographs. \ours verifies the whole candidate pool and is more accurate than the SigLIP retriever on every graph for both models. On the Wikidata graph, whose graph shortlist contains the true species for most queries, verification of the shortlist is more accurate than verification of the pool. The range edge adds a large gain on every iNaturalist graph. In the open-set rows, the true species of each query is removed from the graph. The largest verification score separates these queries from the others better than the largest retrieval similarity on every graph.

\subsection{What a node should show}
\label{sec:modality}
\begin{table}[t]
\centering\footnotesize
\setlength{\tabcolsep}{3.2pt}
\caption{\textbf{Evidence on the node.} Accuracy in percent of pairwise verification of the graph shortlist when the prompt holds the species name, the name and the node text, or the name and the node photograph, where every photograph result differs from the name and text results of the same model with McNemar $p<0.01$.}
\label{tab:modality}
\begin{tabular}{l*{2}{cc>{\columncolor{ourbg}}c}}
\toprule
 & \multicolumn{3}{c}{\qas} & \multicolumn{3}{c}{\qbs} \\
\cmidrule(lr){2-4}\cmidrule(lr){5-7}
Graph & name & text & photo & name & text & photo \\
\midrule
CUB-93 & 42.5 & 45.2 & \best{51.9} & 43.0 & 44.6 & \best{55.4} \\
CUB-200 & 54.6 & 54.0 & \best{63.0} & 49.9 & -- & \best{62.9} \\
Wikidata & 49.2 & 48.7 & \best{72.8} & 51.1 & 52.7 & \best{73.4} \\
iNat birds & 27.1 & -- & \best{30.8} & 25.6 & -- & \best{32.0} \\
iNat fungi & 22.3 & 23.6 & \best{33.1} & 22.6 & 26.4 & \best{34.2} \\
iNat Lepid. & 13.7 & 13.7 & \best{20.5} & 13.6 & 16.0 & \best{23.6} \\
iNat plants & 24.1 & -- & \best{36.0} & 24.3 & -- & \best{38.7} \\
\bottomrule
\end{tabular}
\end{table}

\Cref{tab:modality} varies the evidence on each node under one protocol. The VLM verifies the four candidates of the graph shortlist, and the prompt holds the species name, the name with the node text, or the name with the node photograph. The question in the prompt names the evidence it refers to. The node photograph improves on the name on every graph for both models. The node text stays close to the name, both for Wikipedia descriptions and for expert attribute annotations. The Wikipedia descriptions in \cref{fig:compare} open with size and general shape, which look-alike species share, and the photographs show the plumage that separates them.

\begin{table}[t]
\centering\footnotesize
\setlength{\tabcolsep}{5pt}
\caption{\textbf{Photographs per node.} CUB-93 accuracy in percent of the retriever, which averages the embeddings of the node photographs, and of verification of the graph shortlist, which shows the VLM the node photograph most similar to the query.}
\label{tab:shots}
\begin{tabular}{c c >{\columncolor{ourbg}}c >{\columncolor{ourbg}}c}
\toprule
Photographs & & \multicolumn{2}{c}{Verification} \\
\cmidrule(lr){3-4}
per node & Retriever & \qas & \qbs \\
\midrule
1 & 43.8 & 51.9 & \best{55.4} \\
3 & 65.3 & 66.7 & \best{68.3} \\
30 & \best{85.2} & 70.2 & 73.7 \\
\bottomrule
\end{tabular}
\end{table}

\Cref{tab:shots} varies the number of photographs per node on CUB-93. The retriever profits most from many photographs, because it averages their embeddings into one prototype. Verification compares the query with a single photograph in every setting and keeps more of its accuracy when a node holds one photograph. Wikidata and most catalogues hold one photograph per entity, and the remaining experiments use one photograph per node.

\subsection{Verification and multi-image RAG}
\label{sec:read}
\begin{table}[t]
\centering\footnotesize
\setlength{\tabcolsep}{2.6pt}
\caption{\textbf{Verification and multi-image RAG.} Accuracy of pairwise verification minus that of multi-image RAG in points on the same similarity shortlist of size $K$, over all test queries and over the CUB-200 queries whose shortlist contains the true species, where an asterisk marks McNemar $p<0.05$.}
\label{tab:multi}
\begin{tabular}{lrrrrrr}
\toprule
 & \multicolumn{3}{c}{\qas} & \multicolumn{3}{c}{\qbs} \\
\cmidrule(lr){2-4}\cmidrule(lr){5-7}
Graph & $K{=}4$ & $K{=}8$ & $K{=}12$ & $K{=}4$ & $K{=}8$ & $K{=}12$ \\
\midrule
CUB-200 & $-2.6^{*}$ & $-1.5^{*}$ & $-1.1$ & $-0.8$ & $+0.2$ & $+1.0$ \\
\quad shortlisted only & $+1.0$ & $+0.6$ & $+0.2$ & $+2.7$ & $+2.2$ & $+2.1$ \\
iNat fungi & $+4.5^{*}$ & $+6.9^{*}$ & -- & $+1.3$ & $+4.4^{*}$ & -- \\
iNat Lepidoptera & $+3.7^{*}$ & $+5.8^{*}$ & -- & $+1.7$ & $+1.6$ & -- \\
iNat plants & $+2.9$ & $+2.8$ & -- & $-0.3$ & $+0.3$ & -- \\
\bottomrule
\end{tabular}
\end{table}

Multi-image RAG reads the same node photographs in a single prompt. \Cref{tab:multi} gives both methods identical similarity shortlists. On CUB-200, multi-image RAG is more accurate with \qa at the two smaller shortlists, and the two methods do not differ significantly with \qb. On the CUB-200 queries whose shortlist contains the true species, verification is at least as accurate as multi-image RAG for both models and every shortlist size. Multi-image RAG gains on the remaining queries, where it can still name a species from the full list of the graph. On fungi and Lepidoptera, verification is more accurate at every shortlist size, and every difference for \qa is significant. On plants the two methods do not differ significantly.

\begin{table}[t]
\centering\footnotesize
\setlength{\tabcolsep}{3pt}
\caption{\textbf{Order, variance and cost.} Accuracy in percent of multi-image RAG as the correct photograph moves through the four slots of the prompt, accuracy over three random draws of the node photographs and time per query.}
\label{tab:analysis}
\begin{tabular}{>{\raggedright\arraybackslash}p{5.0cm}cc}
\toprule
 & \qas & \qbs \\
\midrule
\multicolumn{3}{l}{\textit{Multi-image RAG on CUB-93 by slot of the correct photograph}} \\
\quad slot 1 & 86.5 & 74.1 \\
\quad slot 2 & 63.8 & 72.0 \\
\quad slot 3 & 54.6 & 69.9 \\
\quad slot 4 & 57.4 & 67.4 \\
\quad answers naming slot 1 & 43.2 & 31.2 \\
\midrule
\multicolumn{3}{l}{\textit{CUB-200, mean and deviation over three draws of node photographs}} \\
\quad Retriever & 54.3 $\pm$ 1.0 & 54.3 $\pm$ 1.0 \\
\ourrow \quad \ours, pool & 70.0 $\pm$ 1.1 & 69.7 $\pm$ 1.2 \\
\quad Multi-image RAG, $K{=}4$ & 67.3 $\pm$ 0.6 & 66.7 $\pm$ 0.9 \\
\midrule
\multicolumn{3}{l}{\textit{CUB-200, seconds per query on one H100 GPU}} \\
\ourrow \quad \ours, pool & -- & \best{0.67} \\
\quad Multi-image RAG, $K{=}4$ & -- & 0.84 \\
\quad Multi-image RAG, $K{=}8$ & -- & 0.95 \\
\bottomrule
\end{tabular}
\end{table}

\Cref{tab:analysis} examines the order, the variance and the cost of the two readers. Multi-image RAG with \qa answers far more prompts correctly when the correct photograph comes first, and both models name the first slot more often than chance. Retrievers place their best guess in the first slot, and this preference helps multi-image RAG whenever the guess is right. Verification scores each candidate in its own prompt and has no slot order. Redrawing the node photograph of every CUB-200 species changes the accuracy of each method little and keeps the order of the methods in every draw. Verification of the whole pool takes less time per query than multi-image RAG with four photographs, because multi-image RAG writes a justification before its answer.

\subsection{Relations that similarity lacks}
\label{sec:relations}
\begin{table}[t]
\centering\footnotesize
\setlength{\tabcolsep}{2.4pt}
\caption{\textbf{Using the photo location.} Species accuracy in percent on iNaturalist with the same candidate pool, where an asterisk marks McNemar $p<0.05$ against \ours with the range edge and a dagger marks a result without a paired test.}
\label{tab:geo}
\begin{tabular}{>{\raggedright\arraybackslash}p{3.0cm}ccccc}
\toprule
 & \multicolumn{2}{c}{birds} & fungi & Lepid. & plants \\
\cmidrule(lr){2-3}\cmidrule(lr){4-4}\cmidrule(lr){5-5}\cmidrule(lr){6-6}
Method & \qbs & \qas & \qbs & \qbs & \qbs \\
\midrule
Retriever, range edge & 45.6$^{*}$ & 45.6$^{*}$ & 36.4$^{*}$ & 29.1$^{*}$ & 49.5$^{*}$ \\
\ourrow \ours, range edge & 57.6 & \best{58.5} & \best{46.2} & \best{44.9} & \best{58.2} \\
Verification, location in prompt & 56.9 & 52.1$^{*}$ & 43.5 & 43.8 & 57.0 \\
Multi-image, GPS & 40.3$^{*}$ & 36.2$^{*}$ & 33.0$^{*}$ & 25.2$^{*}$ & 42.3$^{\dagger}$ \\
Multi-image, range-ranked, GPS, counts & \best{58.9} & 53.7$^{*}$ & 40.9$^{*}$ & 40.9$^{\dagger}$ & 57.0$^{\dagger}$ \\
Multi-image, fused top four, GPS, counts & 58.5 & 55.4$^{*}$ & -- & -- & -- \\
\bottomrule
\end{tabular}
\end{table}

\paragraph{Range edges.}
A range edge records where a species has been photographed, and an image embedding does not encode this information. \Cref{tab:geo} compares \ours with other uses of the same location on the same candidate pool. The retriever with the range edge stays well below \ours on every graph. Writing the location and the range counts into each verification prompt recovers most of the gain of the fusion. The strongest multi-image prompt ranks the shortlist by the range edge and adds the GPS coordinates and the observation counts of each candidate. That prompt matches \ours on birds with \qb and falls below \ours in the other four columns. Showing the four best candidates of \ours to the multi-image prompt gives no significant gain over \ours.

\begin{figure*}[t]
  \centering
  \includegraphics[width=\textwidth]{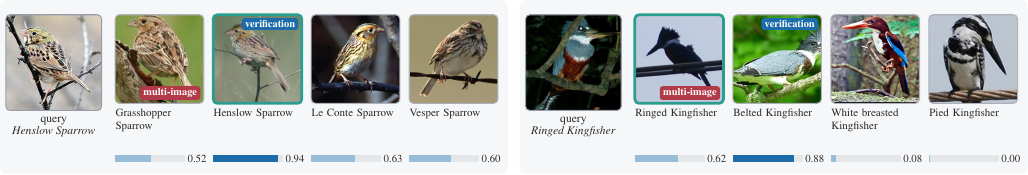}
  \caption{\textbf{Qualitative results.} Graph-shortlist candidates of two CUB-93 queries with the verification scores of \qa, where the frame and the bold name mark the true species and the tags mark the answers of verification and of multi-image RAG.}
  \label{fig:qual}
\end{figure*}
\begin{table}[t]
\centering\footnotesize
\setlength{\tabcolsep}{2.4pt}
\caption{\textbf{Answering from fact edges.} Accuracy in percent over 28{,}820 multiple-choice questions about the diet, migration, habitat, foraging lifestyle and body mass of the bird in a CUB-200 test image, where an asterisk marks McNemar $p<0.05$ against the fact edge of the verified species and a dagger marks a result without a paired test.}
\label{tab:facts}
\begin{tabular}{p{4.1cm}cc}
\toprule
Answer source & \qas & \qbs \\
\midrule
Zero-shot VLM & 47.9$^{*}$ & 44.1$^{*}$ \\
Prompt with facts of the most similar species & 82.2$^{*}$ & 80.5$^{*}$ \\
Prompt with facts of four candidates & 83.3$^{*}$ & 83.2$^{*}$ \\
Prompt with photographs and facts of four candidates & 83.8$^{*}$ & 83.5$^{*}$ \\
Fact edge of the multi-image RAG answer & 88.8$^{\dagger}$ & 87.7$^{\dagger}$ \\
\ourrow \ours, fact edge of the verified species & \best{89.5} & \best{89.2} \\
\gr{Fact edge of the true species} & \gr{100} & \gr{100} \\
\bottomrule
\end{tabular}
\end{table}

\paragraph{Fact edges.}
Fact edges answer questions whose answers the photograph does not show. \Cref{tab:facts} asks five questions per CUB-200 test image and takes the AVONET values of the true species as answers. Prompts that hold the facts of the candidates leave the VLM to pick the species and the fact in one step. \ours separates the two steps and reads the fact edge of the verified species. \ours is significantly more accurate than every prompt for both models. Reading the fact edge of the species that multi-image RAG names is also more accurate than the prompts and stays below \ours.
\begin{table}[t]
\centering\footnotesize
\setlength{\tabcolsep}{2.2pt}
\caption{\textbf{Fact questions by type.} Accuracy in percent on CUB-200 of the prompt with photographs and facts of four candidates and of the fact edge of the species that \ours verifies.}
\label{tab:perq}
\begin{tabular}{llccccc}
\toprule
Model & Source & diet & migration & habitat & lifestyle & mass \\
\midrule
\qas & prompt & 87.1 & 85.9 & 75.0 & 88.1 & 82.9 \\
\ourrow & \ours & \best{91.8} & \best{90.7} & \best{83.0} & \best{93.5} & \best{88.6} \\
\qbs & prompt & 87.6 & 87.2 & 74.9 & 83.1 & 84.6 \\
\ourrow & \ours & \best{91.6} & \best{89.8} & \best{81.9} & \best{93.4} & \best{89.2} \\
\bottomrule
\end{tabular}
\end{table}

\Cref{tab:perq} splits the result by question type, and the fact edge is more accurate than the prompt for every type and both models.

\subsection{Ablation}
\label{sec:ablation}
\begin{table}[t]
\centering\footnotesize
\setlength{\tabcolsep}{2.6pt}
\caption{\textbf{Ablation on CUB-200.} Accuracy in percent of pairwise verification with one sentence on the sibling edge appended to each prompt and with other candidate sets, where an asterisk marks McNemar $p<0.05$ against the graph shortlist without a sentence.}
\label{tab:ablation}
\begin{tabular}{lcccc}
\toprule
Candidates & size & recall & \qas & \qbs \\
\midrule
Graph shortlist & 4 & 78.6 & 63.0 & 62.9 \\
\quad + generic sentence & 4 & 78.6 & -- & 63.6$^{*}$ \\
\quad + note of another species & 4 & 78.6 & 61.8$^{*}$ & 59.5$^{*}$ \\
\ourrow \quad + contrastive note & 4 & 78.6 & 63.1 & 65.1$^{*}$ \\
\midrule
Name group of top species & 8.6 & 83.0 & 62.7 & 62.4 \\
Mined confusion edges & 4 & 83.0 & 67.1 & 66.3 \\
Similarity shortlist & 4 & 80.6 & 65.7 & 65.4 \\
Similarity shortlist & 8 & 89.8 & 67.9 & 67.7 \\
\ourrow Whole pool & 12.8 & 95.1 & \best{69.0} & \best{68.8} \\
\bottomrule
\end{tabular}
\end{table}

\Cref{tab:ablation} varies the prompt and the candidate set on CUB-200. The upper block appends one sentence about the sibling edge to each prompt of the graph shortlist. The contrastive note raises the accuracy of \qb above a generic sentence about the same look-alikes, and the note of another candidate lowers it. The note leaves the accuracy of \qa unchanged. The lower block changes the candidate set. Sibling edges and name groups choose less useful candidates than a similarity shortlist of the same size, and confusion edges mined from retriever errors on the development split improve on that shortlist. Larger similarity shortlists reach higher recall, and verification of the whole pool is the most accurate setting.

\subsection{Qualitative results}
\label{sec:qual}
\Cref{fig:qual} shows two CUB-93 queries with their graph-shortlist candidates. For the Henslow Sparrow, multi-image RAG names the first candidate, and verification gives the true species a clear margin over its look-alikes. For the Ringed Kingfisher, the node photograph of the true species is a backlit silhouette. Verification prefers the Belted Kingfisher in that case, and multi-image RAG answers correctly.

\section{Conclusion}
We presented \ours, a training-free GraphRAG method that stores a real photograph on every node of a knowledge graph and lets a frozen VLM verify the query against the photograph of each candidate. The node photograph gave the VLM evidence that names and descriptions did not give. Pairwise verification of the candidate pool outperformed retrieval, combined with range edges, led to fact edges and detected species missing from the graph. We will release the graphs, prompts and code to support image-grounded GraphRAG beyond species.

{
    \small
    \bibliographystyle{ieeenat_fullname}
    \bibliography{references}
}

\clearpage
\maketitlesupplementary
\setcounter{section}{0}
\renewcommand{\thesection}{\Alph{section}}
\setcounter{figure}{0}\renewcommand{\thefigure}{S\arabic{figure}}
\setcounter{table}{0}\renewcommand{\thetable}{S\arabic{table}}

\section{Fusion form}
\label{sec:supp_fusion}
\begin{table}[h]
\centering\footnotesize
\setlength{\tabcolsep}{3pt}
\caption{\textbf{Fusion form.} Accuracy in percent of an additive form with standardized range scores and of the log-linear form of \cref{eq:fusion}, with weights chosen on the development split.}
\label{tab:supp_fusion}
\begin{tabular}{llcccc}
\toprule
 & & \multicolumn{2}{c}{development} & \multicolumn{2}{c}{test} \\
\cmidrule(lr){3-4}\cmidrule(lr){5-6}
Domain & Model & additive & log-linear & additive & log-linear \\
\midrule
birds & Qwen2.5 & 57.3 & 60.2 & 58.0 & 58.5 \\
birds & Qwen3 & 54.3 & 57.0 & 53.9 & 57.6 \\
fungi & Qwen2.5 & 43.3 & 44.9 & 44.4 & 41.8 \\
fungi & Qwen3 & 43.5 & 44.9 & 40.9 & 46.2 \\
Lepidoptera & Qwen2.5 & 38.9 & 41.9 & 40.8 & 44.4 \\
Lepidoptera & Qwen3 & 38.6 & 41.7 & 41.0 & 44.9 \\
plants & Qwen2.5 & 52.9 & 56.9 & 56.4 & 57.5 \\
plants & Qwen3 & 53.7 & 57.6 & 55.2 & 58.2 \\
\bottomrule
\end{tabular}
\end{table}

\Cref{tab:supp_fusion} compares an additive fusion with the log-linear fusion of \cref{eq:fusion}. The additive form standardizes the retrieval similarity and the range score within the pool and adds them to $p_v$ with weights of at most 2. The development split prefers the log-linear form in all eight settings, and the test split agrees in seven of them.

\section{Evidence on the node}
\begin{table*}[h]
\centering\footnotesize
\setlength{\tabcolsep}{4pt}
\caption{\textbf{Evidence on the node.} Accuracy in percent and pairwise AUC of verification of the graph shortlist for each kind of node evidence, including the node photograph followed by the node text and \qc on CUB-93.}
\label{tab:supp_modality}
\begin{tabular}{llcccc}
\toprule
Graph & Model & name & name and text & name and photograph & name, photograph and text \\
\midrule
CUB-93 & Qwen2.5 & 42.5 / 0.786 & 45.2 / 0.787 & 51.9 / 0.857 & 54.6 / 0.855 \\
CUB-93 & Qwen3 & 43.0 / 0.751 & 44.6 / 0.769 & 55.4 / 0.850 & 57.5 / 0.866 \\
CUB-93 & Qwen3-32B & 50.0 / 0.844 & 50.0 / 0.841 & 57.8 / 0.901 & -- \\
CUB-200 & Qwen2.5 & 54.6 / 0.851 & 54.0 / 0.841 & 63.0 / 0.913 & -- \\
CUB-200 & Qwen3 & 49.9 / 0.822 & -- & 62.9 / 0.910 & -- \\
Wikidata & Qwen2.5 & 49.2 / 0.776 & 48.7 / 0.750 & 72.8 / 0.884 & 71.5 / 0.869 \\
Wikidata & Qwen3 & 51.1 / 0.741 & 52.7 / 0.784 & 73.4 / 0.877 & 73.9 / 0.890 \\
iNat birds & Qwen2.5 & 27.1 / 0.792 & -- & 30.8 / 0.802 & -- \\
iNat birds & Qwen3 & 25.6 / 0.773 & -- & 32.0 / 0.827 & -- \\
iNat fungi & Qwen2.5 & 22.3 / 0.702 & 23.6 / 0.711 & 33.1 / 0.816 & -- \\
iNat fungi & Qwen3 & 22.6 / 0.719 & 26.4 / 0.768 & 34.2 / 0.838 & -- \\
iNat Lepid. & Qwen2.5 & 13.7 / 0.695 & 13.7 / 0.665 & 20.5 / 0.803 & -- \\
iNat Lepid. & Qwen3 & 13.6 / 0.672 & 16.0 / 0.695 & 23.6 / 0.845 & -- \\
iNat plants & Qwen2.5 & 24.1 / 0.678 & -- & 36.0 / 0.760 & -- \\
iNat plants & Qwen3 & 24.3 / 0.690 & -- & 38.7 / 0.816 & -- \\
\bottomrule
\end{tabular}
\end{table*}

\Cref{tab:supp_modality} adds pairwise AUC to \cref{tab:modality}, which measures how well $p_v$ separates candidates of the true species from other candidates. The table also reports the node photograph followed by the node text and the results of \qc on CUB-93.

\section{Notes on further graphs}
\begin{table*}[h]
\centering\footnotesize
\setlength{\tabcolsep}{3.5pt}
\caption{\textbf{Notes on further graphs.} Accuracy in percent of verification of the graph shortlist with a sentence on the sibling edge and the McNemar $p$ against no sentence, where the LLM notes are written by Qwen2.5-14B~\cite{qwen25llm2024} from Wikipedia text.}
\label{tab:supp_notes}
\begin{tabular}{llcccccc}
\toprule
Graph & Model & no sentence & attribute note & generic & note of another & LLM note & LLM note of another \\
\midrule
CUB-93 & Qwen2.5 & 51.9 & 54.8 ($p{=}0.071$) & 53.2 ($p{=}0.4$) & 54.0 ($p{=}0.26$) & 52.4 ($p{=}0.88$) & -- \\
CUB-93 & Qwen3 & 55.4 & 59.1 ($p{=}0.016$) & -- & 50.8 ($p{=}0.024$) & 56.2 ($p{=}0.7$) & -- \\
CUB-93 & Qwen3-32B & 57.8 & 58.6 ($p{=}0.68$) & -- & -- & -- & -- \\
iNat fungi & Qwen2.5 & 33.1 & -- & -- & -- & 30.2 ($p{=}0.0045$) & 31.4 ($p{=}0.13$) \\
iNat fungi & Qwen3 & 34.2 & -- & -- & -- & 34.6 ($p{=}0.74$) & 33.1 ($p{=}0.42$) \\
iNat Lepidoptera & Qwen2.5 & 20.5 & -- & -- & -- & 19.2 ($p{=}0.18$) & 18.7 ($p{=}0.063$) \\
iNat Lepidoptera & Qwen3 & 23.6 & -- & -- & -- & 23.5 ($p{=}1$) & 21.9 ($p{=}0.07$) \\
\bottomrule
\end{tabular}
\end{table*}

\Cref{tab:supp_notes} repeats the note experiment of \cref{tab:ablation} on CUB-93 with three models and tests notes that an LLM writes from Wikipedia text on iNat fungi and Lepidoptera. A further CUB-93 variant crops the body part that the note names and adds the crop to the prompt of \qa. That variant reaches 52.2\%, against 51.6\% for a random body part and 51.9\% without a crop.

\section{Multi-image RAG}
\begin{table}[h]
\centering\footnotesize
\setlength{\tabcolsep}{3pt}
\caption{\textbf{CUB-93 shortlisted queries.} Accuracy in percent on the CUB-93 queries whose graph shortlist contains the true species.}
\label{tab:supp_multi}
\begin{tabular}{lcc}
\toprule
 & \qas & \qbs \\
\midrule
Multi-image RAG & 73.0 & 72.7 \\
Verification & 68.8 & 73.0 \\
Graph-shortlist recall & 75.8 & 75.8 \\
\bottomrule
\end{tabular}
\end{table}

\begin{table}[h]
\centering\footnotesize
\setlength{\tabcolsep}{3pt}
\caption{\textbf{Chaining.} Accuracy in percent of verification of the pool and of showing its four highest-scored candidates to the multi-image prompt, where an asterisk marks McNemar $p<0.05$.}
\label{tab:supp_chain}
\begin{tabular}{lcccc}
\toprule
 & \multicolumn{2}{c}{\qas} & \multicolumn{2}{c}{\qbs} \\
\cmidrule(lr){2-3}\cmidrule(lr){4-5}
Graph & pool & chained & pool & chained \\
\midrule
CUB-93 & 59.7 & 63.4$^{*}$ & 61.8 & 58.3 \\
CUB-200 & 69.0 & 70.4$^{*}$ & 68.8 & 68.3 \\
Wikidata graph & 70.7 & 71.0 & 69.4 & 68.3 \\
iNat fungi & 34.5 & 30.9$^{*}$ & 38.1 & 35.2 \\
iNat Lepidoptera & 28.4 & 27.5 & 32.6 & 31.2 \\
iNat plants & 38.8 & 36.9 & 42.6 & 43.0 \\
\bottomrule
\end{tabular}
\end{table}

On CUB-93, multi-image RAG is more accurate than verification also on the queries whose graph shortlist contains the true species, as \cref{tab:supp_multi} shows. \Cref{tab:supp_chain} verifies the pool and shows its four highest-scored candidates to the multi-image prompt.

\section{Weighted fact edges}
\begin{table}[h]
\centering\footnotesize
\setlength{\tabcolsep}{1.8pt}
\caption{\textbf{Weighted fact edges.} Accuracy in percent on CUB-200 when the answers of all pool candidates are weighted by $\exp(6\,(p_v+\lambda\tilde\sigma_v))$, with the share of queries whose pool candidates share one answer.}
\label{tab:supp_perq}
\begin{tabular}{l>{\raggedright\arraybackslash}p{1.9cm}ccccc}
\toprule
Model & Source & diet & migration & habitat & lifestyle & mass \\
\midrule
\qas & weighted fact edges & 91.5 & 90.9 & 82.2 & 93.6 & 88.4 \\
\qbs & weighted fact edges & 91.0 & 89.7 & 80.9 & 92.2 & 88.2 \\
Both & candidates agree & 29.0 & 14.3 & 2.0 & 15.6 & 1.0 \\
\bottomrule
\end{tabular}
\end{table}

A weighted variant of fact answering gives every pool candidate the weight $\exp(6\,(p_v+\lambda\tilde\sigma_v))$, sums the weights of the candidates that share an answer and returns the answer with the largest total. \Cref{tab:supp_perq} reports this variant by question type.

\section{\qc on CUB-93}
\begin{table}[h]
\centering\footnotesize
\setlength{\tabcolsep}{3pt}
\caption{\textbf{\qc on CUB-93.} Species accuracy in percent with one photograph per node.}
\label{tab:supp_32b}
\begin{tabular}{lc}
\toprule
Method & accuracy \\
\midrule
Retriever & 43.8 \\
Verification of the graph shortlist & 57.8 \\
Verification of the pool & 63.4 \\
Multi-image RAG on the graph shortlist & 62.9 \\
\quad on queries whose shortlist contains the true species & 78.0 \\
Verification on the same queries & 76.6 \\
\bottomrule
\end{tabular}
\end{table}

\Cref{tab:supp_32b} collects the results of \qc on CUB-93.

\section{Prompts}
The verification prompt has a system message and a user message. The system message reads ``You are an expert ornithologist.'' The user message shows ``Image 1 (the bird to identify):'' with the query, ``Image 2 (a reference photo of a \{name\}):'' with the node photograph and the question ``Is the bird in Image 1 the same species as the reference bird in Image 2 (\{name\})? Compare them carefully. Answer with Yes or No only.'' For the evidence ablation the photograph is removed or followed by ``Expert attribute description of the \{name\}: \dots'', and the question becomes ``Is the bird in Image 1 a \{name\}?'' for the name alone and ``Does the bird in Image 1 match this description of the \{name\}, i.e. is it a \{name\}?'' for the text. For the notes ``Field-guide note: compared with its look-alikes (\dots), the \{name\} \dots'' is inserted before the question. On iNaturalist the word bird is replaced by fungus, insect or plant. Multi-image RAG shows the query and then ``Reference $i$ (labeled: \{name\})'' with each photograph, asks the model to compare the query with each reference and to finish with ``Answer:'' followed by a species from the full list. Fact questions are multiple-choice questions over the values in the graph, and the answer letter is read from the logits.

\end{document}